\documentclass{article}

\usepackage{microtype}
\usepackage{graphicx}
\usepackage{subfigure}
\usepackage{amsfonts}
\usepackage{booktabs} 

\usepackage{hyperref,steinmetz}

\usepackage[accepted]{icml2018}

\usepackage{url}
\usepackage{color}

\icmltitlerunning{Scale equivariance in CNNs with vector fields}

\begin{document}

\twocolumn[
\icmltitle{Scale equivariance in CNNs with vector fields}



\icmlsetsymbol{equal}{*}

\begin{icmlauthorlist}
\icmlauthor{Diego Marcos}{wur}
\icmlauthor{Benjamin Kellenberger}{wur}
\icmlauthor{Sylvain Lobry}{wur}
\icmlauthor{Devis Tuia}{wur}
\end{icmlauthorlist}

\icmlaffiliation{wur}{Wageningen University, Netherlands}

\icmlcorrespondingauthor{Diego Marcos}{diego.marcos@wur.nl}


\vskip 0.3in
]



\printAffiliationsAndNotice{}  

\begin{abstract}
We study the effect of injecting local scale equivariance into Convolutional Neural Networks. This is done by applying each convolutional filter at multiple scales. The output is a vector field encoding for the maximally activating scale and the scale itself, which is further processed by the following convolutional layers. This allows all the intermediate representations to be locally scale equivariant. We show that this improves the performance of the model by over $20\%$ in the scale equivariant task of regressing the scaling factor applied to randomly scaled MNIST digits. Furthermore, we find it also useful for scale \emph{invariant} tasks, such as the actual classification of randomly scaled digits. This highlights the usefulness of allowing for a compact representation that can also learn relationships between different local scales by keeping internal scale equivariance.
\end{abstract}

\section{Introduction}

Equivariance to a predefined set of transformations can be a very desirable property of computer vision models. For instance, translation equivariance has a big role in the success of Convolutional Neural Networks (CNN) when applied to images, since in many applications a pattern on the image should be detected independently from its location. This can be clearly seen in problems such as semantic segmentation or optical flow estimation, where a translation of the input image should results in a translation of the output.


In this work, we propose a method for enforcing scale equivariance within CNNs. Our proposition has three main steps:

\noindent 1) A convolutional filter is applied at multiple scales and only information about the maximally activating scale at each location is passed to the output.

\noindent 2) The output is composed of the magnitude of the maximally activating scale \emph{and} of the corresponding scale value itself. These two values are combined into a 2D vector.

\noindent 3) Convolutions on this vector field use a similarity metric that takes into account both magnitudes and scale factors.

We propose an end-to-end learnable framework based on these principles that can be applied to any kind of convolutional architecture and allow encoding both invariance and equivariance to scale, depending on the nature of the task. In this paper, we validate the idea on a scale-invariant image classification problem, where digits from the MNIST dataset are randomly scaled, and on a scale equivariant regression problem, where we aim at regressing the scaling factor applied to the digits. In both cases, we outperform state of the art scale-invariant models.

\begin{figure*}[h]
\vskip 0.2in
\begin{center}
\centerline{\includegraphics[width=\textwidth]{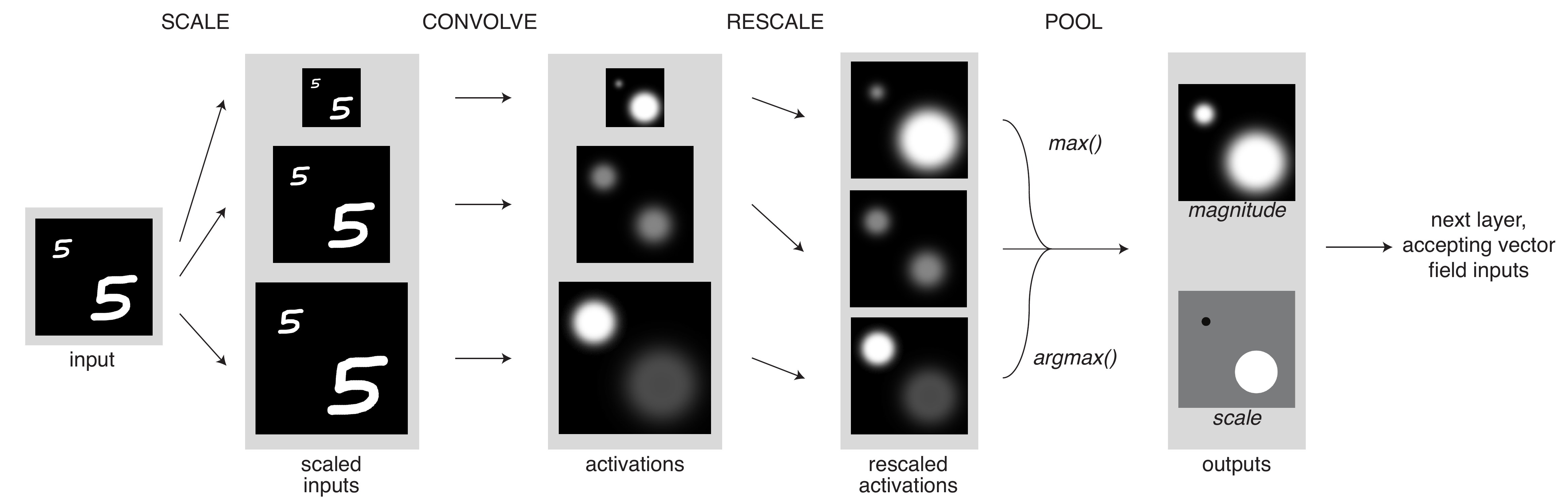}}
\vskip -0.1in
\caption{Working principle of the proposed scale-equivariant CNN. We convolve scaled versions of the input with identical filters and rescale the activations to equal size. The output is then composed of max-pooled activation maps, together with the scale per location (argmax). In the example above, the layer will detect both numbers in the image (scale invariance) and will also provide the information at which scale the fives were detected, allowing for scale equivariance.}
\label{fig:scaleEq_principle}
\end{center}
\vskip -0.2in
\end{figure*}

\section{Related Work}

Our model builds on the work on local scale invariance by~\cite{kanazawa2014locally}, where filters are applied at multiple scales and only the maximum activation value per location is kept, as well as on an application to scale of the vector field representation of~\cite{Marcos_2017_ICCV}, that was proposed to disentangle rotation and content information.

\begin{figure*}[t]
\vskip 0.2in
\begin{center}
\centerline{\includegraphics[width=1.9  \columnwidth]{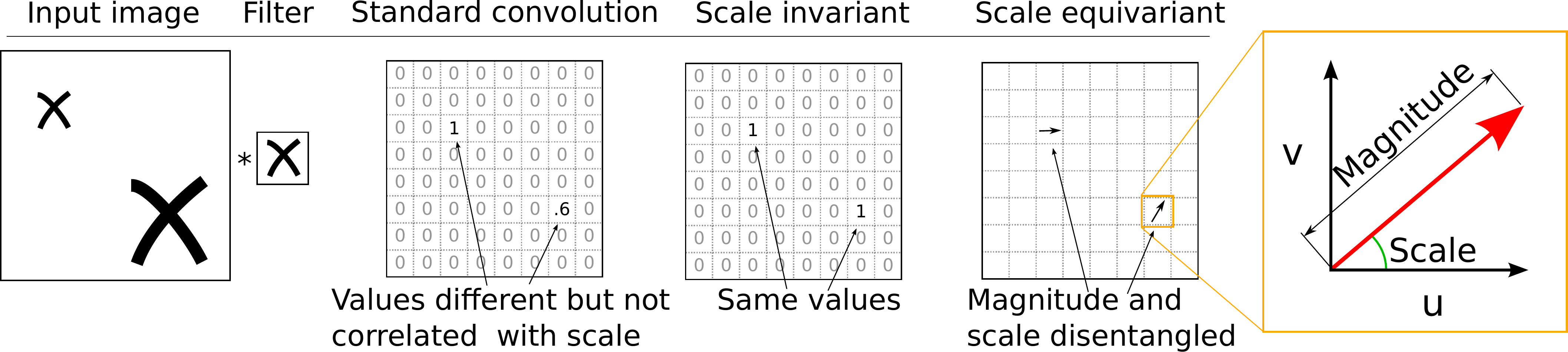}}
\caption{Given two input images, where patterns appear at different relative scales, and a convolutional filter learned to detect this pattern (left), a locally scale invariant model is designed to provide two indistinguishable results (middle), while a locally scale equivariant model is built to disentangle the \emph{amount} of presence of the pattern and the \emph{scale} at which it is most present. In this case these two values are represented as the \emph{magnitude} and \emph{orientation} of the output vectors (right).}
\label{fig:eq_vs_inv}
\end{center}
\vskip -0.3in
\end{figure*}

The modified convolution operator presented by~\cite{kanazawa2014locally} is locally scale invariant and consists of all the elements shown in Fig.~\ref{fig:scaleEq_principle}, except that the output consists only of the magnitude map. If we apply it to an image and look at the value at a single location of the output, small variations in scale of the underlying local object will have no effect on this value (see Fig.~\ref{fig:eq_vs_inv} middle). This is precisely the sought behavior. However, if these scale variations contained information useful for the task at hand but were discarded, we would be hampering the model's performance. For instance, in histology and cytology, the sizes of cells and organelles can provide hints about their functionality, while in scene understanding, apparent object sizes implicitly encode distance from the camera and can be valuable information for scene segmentation~\cite{zhang2017scale}.

Other methods in the literature can be used to obtain global (but not local) scale invariance, and consist of applying multiple Siamese networks with either differently scaled filters~\cite{xu2014scale} or differently scaled input images~\cite{laptev2016ti}. As noted by the authors of~\cite{laptev2016ti}, enforcing scale invariance can lead to a loss in performance. This might happen when the relative sizes of certain features on the image are important for the task: suppose we want a model that detects whether an image contains a duck family. A scale-invariant duck detector with a single appearance model will simply detect multiple ducks, aggravating the task of distinguishing a duck family from a flock of similarly sized ducks. Making equivariance explicit in the model can disentangle appearance and scale, allowing to keep a single appearance model. This maintains the advantages of scale-invariant models by still allowing a single appearance model, while retaining the information about relative sizes of the detected elements. Such a model will be able to distinguish the presence of a larger duck accompanied by a number of ducklings. 

Equivariance to other transformations, chiefly rotation, through weight sharing has gathered a lot of attention in recent literature~\cite{cohen2016group,dieleman2016exploiting,ravanbakhsh2017equivariance,weiler2017learning,zhou2017oriented}. Some works propose to use enriched representations (beyond scalar fields) encoding equivariance while maintaining more compact intermediate representations, based on vector, tensor or complex number fields~\cite{Marcos_2017_ICCV,thomas2018tensor,worrall2017harmonic}.
In this contribution, we propose to use a vector field representation similar to the one in~\cite{Marcos_2017_ICCV} to obtain a locally scale-equivariant deep CNN.

\section{Scale equivariance}
Let $\mathbf{x}\in \mathcal{X} \equiv \mathbb{R}^{M\times N}$ be an image containing an object or pattern of interest and $\mathbf{y}=f(\mathbf{x}) \in \mathcal{Y} \equiv \mathbb{R}^{M\times N \times d}$ be the scalar ($d=1$) or tensorial result of applying the feature extractor $f$ to $\mathbf{x}$. Let us define a group of transformations $G$. A representation $T$ of $G$ is a function from $G$ to a family of linear operators such that, for $g,h\in G$:
\begin{equation}
T(g)T(h) = T(gh).
\end{equation}
In our case, let $g$ correspond to the scaling of an object or pattern and $T^\mathcal{X}(g)$ to a linear operator that implements this scaling, such as through some interpolation method. We say that $f$ is scale \emph{invariant} if:
\begin{equation}
f(T^\mathcal{X}(g) \mathbf{x}) = f(\mathbf{x}),
\end{equation}
while we say that $f$ is scale \emph{equivariant} if we can find some representation $T^\mathcal{Y}$ of $G$ on $\mathcal{Y}$, other than the identity, such that:
\begin{equation}
f(T^\mathcal{X}(g) \mathbf{x}) = T^\mathcal{Y}(g)f(\mathbf{x}).
\end{equation}

In this work $\mathbf{y}\in\mathcal{Y}\equiv \mathbb{R}^{M\times N \times 2}$ is a vector field where scale is encoded as the \emph{angle} $\theta$ of the vectors in polar coordinates:
\begin{equation}
\mathbf{y} = \mathbf{\rho} \, \phase{\mathbf{\theta}},
\end{equation}

where $\rho \in \mathbb{R}^{M\times N}$ is map of magnitudes of feature responses and $\theta \in \mathbb{R}^{M\times N}$ a map of values that depend linearly on the scale of these responses. This allows to define:
\begin{equation}
T^\mathcal{Y}(g_s)\mathbf{y} := T^\mathcal{X}(g_s)\mathbf{\rho}, \phase{T^\mathcal{X}(g_s)\mathbf{\theta} + s},
\end{equation}
where the phase changes proportionally to the scaling factor.

\section{Scale-Equivariant convolution with vector fields}

We propose to build a scale equivariant CNN by using convolutions that apply each learned filter at different scales and then \emph{pool across scales}, as shown in Fig.~\ref{fig:scaleEq_principle}. Similarly to~\cite{kanazawa2014locally}, we obtain multiple copies of the input image or tensor using bilinear interpolation. We then apply a convolution with the same filter to this set of inputs, interpolate the outputs back to the original size and perform a max-pooling operation across the different scales.
After such scale-pooling, we enrich the output of the modified convolution (the largest magnitude) with the \emph{argmax} across scales, \emph{i.e.} information about the scale that most activated the filter at that location. Fig.~\ref{fig:eq_vs_inv}~(right) exemplifies the equivariance to local scaling provided by this operator.
A convolution filter applied on this output also has to be formed by elements containing magnitude and scale. It is straightforward to see that the na\"ive use of the dot product as a similarity metric would not provide the desired results: we want the similarity to be high when the scales in both the input and the filter are close to each other, and not just when both values are high. Although it would be worth investigating the use of distance based metrics, in this work we follow an approach comparable to~\cite{Marcos_2017_ICCV} and encode the two values as a 2D vector in Cartesian coordinates where the length is the magnitude of the max-pooling and the angle is proportional to the argmax. The vector field convolution operator we use is the same as described in~\cite{Marcos_2017_ICCV} and is based on applying two standard convolutions, corresponding to each orthogonal component in the vector fields, and summing both results.

The total range of angles from the minimum to the maximum scale determines the interaction between different scales in the input and the filter. If the angle range is smaller than $90^o$, there will be a certain level of positive interaction between the smallest and the largest scales, while a range larger than $90^o$ means that the smallest and the largest scales interact negatively in terms of similarity. The range should not surpass $180^o$. A larger range would mean that the smallest and the largest scales would not be the most dissimilar.

\section{Experiments}

In order to showcase the potential of the proposed model, we apply it to the task of simultaneous classification and scale factor regression.

\subsection{MNIST-scale}

MNIST-\emph{scale} is a variation of the MNIST digit classification benchmark introduced by~\cite{sohn2012learning}. It is built by rescaling each image in the original dataset by a factor randomly sampled from a uniform distribution between 0.3 and 1, followed by zero-padding to a size of $28\times 28$ pixels.
We randomly rescale all the images in the original dataset and randomly select 10k samples for training, 2k for validation and 50k for testing. All the results shown are averages over six realizations of the modified dataset.

\subsection{Architecture}

In all our experiments, we use an architecture with three convolutional layers, all with $7\times 7$ filters. The scale invariant and equivariant architectures use 12, 32 and 48 filters in each layer, while the standard CNN was found to provide the best results with three times as many filters. In all models, the last convolutional block is followed by a fully connected layer with 256 hidden units that is then mapped to predict the class, as well as the scale factor in the standard and the invariant models. The regression is based on an MSE loss. In the scale equivariant model, only the magnitudes of the 48-dimensional representation are used for the fully connected layer, which is then solely used to predict the class. The scale factor is directly predicted by linearly combining the angles in the 48-dimensional representation, without an additional hidden layer. We used a total of eight scales, three larger and four smaller than the original, using a scale factor of 1.25 between them. The vector representation was empirically chosen to have an angle range of $120^o$.

\section{Results and discussion}

\noindent \textbf{MNIST-scale classification}. Table~\ref{tab:classif} shows the results of the three tested models, compared to previously published results on the same dataset. \cite{kanazawa2014locally} observed a $0.35\%$ improvement by making a two-layer CNN locally invariant to scale. Similarly, we obtain a $0.38\%$ improvement by making a three-layers CNN locally scale-invariant. Interestingly, an additional $0.31\%$ can be obtained by switching from local invariance to local equivariance, \emph{i.e.} moving from our model using only magnitude to the full model using 2D vector fields. This means that, even for a scale invariant task like MNIST-\emph{scale} classification, keeping the information about the selected scales, thus allowing the model to learn the interactions between different relative scales, can add a substantial boost in performance.

\vspace{-0.2 cm}

\begin{table}[h]
\caption{Classification errors on the MNIST-\emph{scale} 50k test set. Averages and standard deviations over 6 folds.}
\label{tab:classif}
\vskip 0.15in
\begin{center}
\begin{tabular}{lc}
\toprule
Method & Class. error\\
\midrule
TIRBM~\cite{sohn2012learning} & 5.5\\
CNN~\cite{kanazawa2014locally} & 3.48 $\pm$ 0.23\\
SI-CNN~\cite{kanazawa2014locally} & 3.13 $\pm$ 0.19\\
Standard 3-layer CNN  & 3.13 $\pm$ 0.11\\
\midrule
Scale invariant 3-layer CNN  & 2.75 $\pm$ 0.09\\
Scale equivariant 3-layer CNN  & \textbf{2.44} $\pm$ 0.07\\
\bottomrule
\end{tabular}
\end{center}
\vskip -0.1in
\end{table}

\noindent \textbf{MNIST-scale scale factor regression}. The results on the scale factor regression are shown in Table~\ref{tab:regress}. In this case, we observe no improvement at all from injecting scale invariance into the model, and even a slight decrease in accuracy. This is to be expected, since the scale-invariant model explicitly removes information on scale, potentially hampering the regression task. On the other hand, there is a substantial improvement in the scale factor prediction by using the  scale-equivariant model, since the orientation of the vectors in the vector field layers is built to be linearly dependent on the scale of the features found in the input image.

\vspace{-0.2 cm}

\begin{table}[h]
\caption{RMSE of the scale factor regression on the MNIST-\emph{scale} 50k test set. Averages and standard deviations over 6 folds.}
\label{tab:regress}
\vskip 0.15in
\begin{center}
\begin{tabular}{lc}
\toprule
Method & Scale RMSE \\
\midrule
Standard CNN  & 0.254$\pm$ 0.0028\\
Scale invariant CNN  &  0.256$\pm$ 0.0024\\
Scale equivariant CNN  & \textbf{0.206}$\pm$ 0.0017\\
\bottomrule
\end{tabular}
\end{center}
\vskip -0.1in
\end{table}

\vspace{-0.2 cm}
\section{Conclusion}

We presented a method for injecting equivariance to local scale variations in images. This is done by applying each convolutional filter at multiple scales and outputting, for each location, the magnitude and the scale value of the maximally activated scale. These two values are represented as a 2D vector whose length encodes the magnitude and whose orientation encodes the scale. A vector field convolution can be applied on this output to build a deep architecture.

The results on MNIST-\emph{scale} show that a locally scale equivariant model provides a substantial improvement over a scale-invariant one even in a task that is itself scale-invariant by nature and a 3-fold reduction in the number of learnable filters with respect to the best standard CNN model. This highlights the importance of allowing the model to learn the relationships between different relative scales. Less surprisingly, we also found that scale equivariance helps in the task of predicting the scale factor that has been applied to an object in the input image.

\bibliography{example_paper}
\bibliographystyle{icml2018}

\end{document}